%% file: main.tex
\documentclass[10pt,twocolumn,letterpaper]{article}

\usepackage[pagenumbers]{cvpr} 

\input{preamble}

\definecolor{cvprblue}{rgb}{0.21,0.49,0.74}
\usepackage[pagebackref,breaklinks,colorlinks,allcolors=cvprblue]{hyperref}

\def\paperID{} 
\def\confName{CVPR}
\def\confYear{2026}

\title{In-Video Instructions: Visual Signals as Generative Control}

\author{
\bf Gongfan Fang \quad Xinyin Ma \quad Xinchao Wang$^{\dag}$\\
{National University of Singapore} \\
{\tt\small \{gongfan,maxinyin\}@u.nus.edu, xinchao@nus.edu.sg} \\
{\small\textsuperscript{\dag}Corresponding author} \\
{\small Project Page: \url{https://fangggf.github.io/In-Video/}} \\
}

\begin{document}
\twocolumn[{%
\renewcommand\twocolumn[1][]{#1}%
\maketitle
\begin{center}
    \centering
    \captionsetup{type=figure}
    \vspace{-28pt}
    \includegraphics[width=\textwidth]{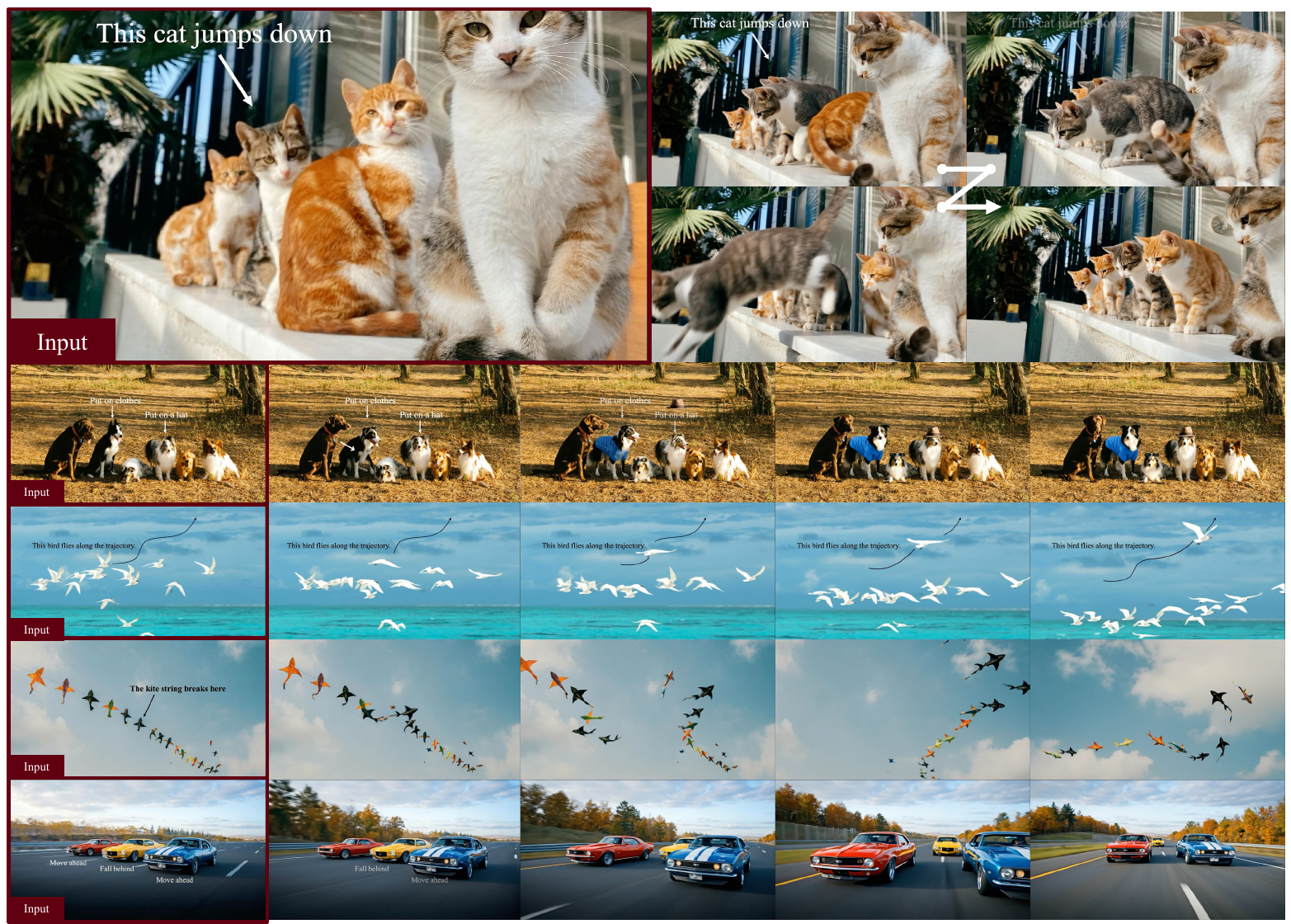}
    \vspace{-18pt}
    \caption{
    \label{fig:teaser}
    Videos generated with the proposed In-Video Instructions. The textual prompt is fixed as ``Follow the instructions step by step,'' while the model synthesizes content purely from the embedded visual signals within the input frames. Zoom in for more details.} 
\end{center}%
}]

\begin{abstract}
Large-scale video generative models have recently demonstrated strong visual capabilities, enabling the prediction of future frames that adhere to the logical and physical cues in the current observation. In this work, we investigate whether such capabilities can be harnessed for controllable image-to-video generation by interpreting visual signals embedded within the frames as instructions, a paradigm we term In-Video Instruction. In contrast to prompt-based control, which provides textual descriptions that are inherently global and coarse, In-Video Instruction encodes user guidance directly into the visual domain through elements such as overlaid text, arrows, or trajectories. This enables explicit, spatial-aware, and unambiguous correspondences between visual subjects and their intended actions by assigning distinct instructions to different objects. Extensive experiments on three state-of-the-art generators, including Veo 3.1, Kling 2.5, and Wan 2.2, show that video models can reliably interpret and execute such visually embedded instructions, particularly in complex multi-object scenarios. 
\end{abstract}


\section{Introduction}
\label{sec:intro}

Large-scale video generative models have recently demonstrated remarkable capabilities in visual understanding, reasoning, and physical-world simulation~\cite{wiedemer2025video,wang2024worlddreamer,li2025worldmodelbench,xiang2024pandora,huang2025vid2world,ren2025videoworld}. These abilities enable models to synthesize temporally coherent and logically consistent video content conditioned on the contextual information present in the current frame. A growing body of recent work highlights this potential across diverse domains, including visual perception~\cite{wiedemer2025video}, manipulation~\cite{li2025novaflow}, puzzle solving~\cite{liu2024solving}, and mathematical reasoning~\cite{wiedemer2025video,tong2025thinking}. 

Such visual ability naturally raises an intriguing question: if a video generative model can interpret visual signals to predict future dynamics, can those same signals also act as an internal control mechanism for video generation in a zero-shot manner? Compared to conventional textual prompts, which provide only coarse descriptions of the intended content, we examine a setting for image-to-video generation in which human guidance is embedded directly into the first video frame. The guidance is conveyed through visual elements such as overlaid text, arrows, trajectories, or other simple markers as shown in Figure~\ref{fig:teaser}. This formulation introduces an additional spatial dimension of control, allowing instructions to be placed near the target objects or regions, and enabling arrows to specify the intended direction or area of influence. These visual instructions, embedded as part of the video itself, provide fine-grained and unambiguous guidance. By interpreting these visual signals, the model is expected to produce the desired behaviors, including plausible object motion, coherent interactions, and precise localization.

Building on these motivations, we introduce \textbf{In-Video Instruction}, a paradigm that encodes user intent directly within the visual input and enables video generative models to interpret this intent as part of the scene semantics. Our method adopts an extremely simple design composed of two basic elements: short textual commands and arrows. The textual commands describe the intended behavior of an object, such as motion or interaction, while the arrows serve as spatial indicators that localize the target or specify the interaction direction. This formulation enables zero-shot and flexible controllability without any retraining. A key advantage of the paradigm is its natural compatibility with complex scenarios, including scenes containing multiple objects and tasks requiring multi-step actions. With instructions grounded in the visual space, different objects can be guided independently, and their behaviors can be specified through multiple sequential or independent instructions. These properties make In-Video Instruction an expressive interface for controllable video generation.

Accordingly, we validate this paradigm across several state-of-the-art video generative models, including both proprietary models such as Veo~3.1~\cite{veo2025}, Kling~2.5~\cite{Kuaishou2024KlingAI}, and open-source models like Wan~2.2~\cite{wan2025wan}.
Our experiments examine a wide range of capabilities, evaluating whether models can (1) comprehend and execute text embedded within visual inputs, (2) accurately localize and associate instructions with specific subjects, (3) generate fine-grained object and camera motions, and (4) follow multiple sequential or independent instructions in complex scenes. The results show that In-Video Instructions offer a clear advantage in tasks that rely on spatial grounding. Models can more reliably bind instructions to the correct subjects and resolve object-specific behaviors, especially in multi-object or cluttered scenes.

In summary, In-Video Instruction provides a direct and flexible interface for expressing user intent within the visual domain. By embedding guidance into the input itself, the paradigm allows video models to interpret instructions through the same mechanisms used for perception, enabling precise, interpretable, and spatially aligned control. 

\section{Related Works}

\paragraph{Video Models as Zero-shot Reasoner.} Large-scale video generative models~\cite{brooks2024video,veo2025,wan2025wan,Kuaishou2024KlingAI,polyak2024movie,seawead2025seaweed,kong2024hunyuanvideo,genmo2024mochi} have recently demonstrated impressive capabilities in understanding and reasoning, enabling them to perform perception, physical modeling, manipulation, and reasoning tasks through video generation~\cite{wiedemer2025video,tong2025thinking,kang2024far,fei2024video}. At the core of these abilities lies the understanding of the current frame’s content and the generation of subsequent frames that follow coherent physical and semantic rules. Recent studies have further investigated these emerging capabilities, examining their generalization to visual puzzle-solving and mathematical reasoning~\cite{tong2025thinking}, physical manipulation~\cite{li2025novaflow}, autonomous driving~\cite{wang2025mila}, and domain-specific knowledge in medical applications~\cite{lai2025video}. Notably, large-scale video generative models have demonstrated the ability to understand textual and symbolic information embedded within videos~\cite{wiedemer2025video,tong2025thinking}. Building upon this capability, this work explores whether such understanding can be leveraged to control video generation, and enable the model to follow instructions embedded directly within the video itself, rather than relying on external textual prompts.

\paragraph{Controllable Video Generation.} 
Recent advancements in video generative models have spurred increasing interest in controllable video generation, which seeks to synthesize videos that accurately reflect user intent~\cite{ma2025controllable,wang2025unianimate,peng2024controlnext}. Early approaches predominantly relied on text-to-video models~\cite{yang2024cogvideox,hong2022cogvideo,jiang2024videobooth}, where high-dimensional visual content is generated from low-dimensional textual descriptions. However, such text-only conditioning often fails to convey complex spatiotemporal semantics. To overcome this limitation, recent studies have incorporated a variety of non-textual modalities, including initial frames for image-to-video generation~\cite{yang2025versatile}, depth maps~\cite{hu2025-DepthCrafter}, canny edges~\cite{zhang2305controlvideo}, bounding boxes~\cite{wang2024boximator}, trajectories~\cite{geng2024motionprompting}, 3D condition~\cite{gu2025das}, sketch~\cite{wang2023videocomposer} and motions~\cite{li2025magicmotion,xiao2024trajectory,hou2024training,xiao2024video,wu2025any2caption,burgert2025go}, thereby enabling fine-grained and multimodal control over the generation process. Moving beyond single-condition control, emerging frameworks aim for multi-condition controllable generation~\cite{wang2025unianimate,xi2025omnivdiff,jiang2025vace}, jointly leveraging visual, spatial, and temporal information to enhance compositional reasoning and creative flexibility.

\section{In-Video Instructions}

This section introduces~\emph{In-Video Instruction}, a controllable video generation paradigm that embeds human intent directly into the visual input, as shown in Figure~\ref{fig:framework}. In contrast to conventional prompt–based conditioning, which requires the model to infer object identity and spatial relations from language alone, In-Video Instruction establishes explicit correspondences between visual subjects and their associated commands by placing the instructions inside the frame. Each frame functions as an interactive canvas where guidance is overlaid as simple visual elements. The pretrained video generative model jointly interprets these elements together with the underlying scene, enabling fine-grained and spatial-aware control over motion and interaction without any retraining or architectural modification.

\subsection{Embedding Instructions into Video Frames}

\begin{figure}[t]
    \centering
    \includegraphics[width=\linewidth]{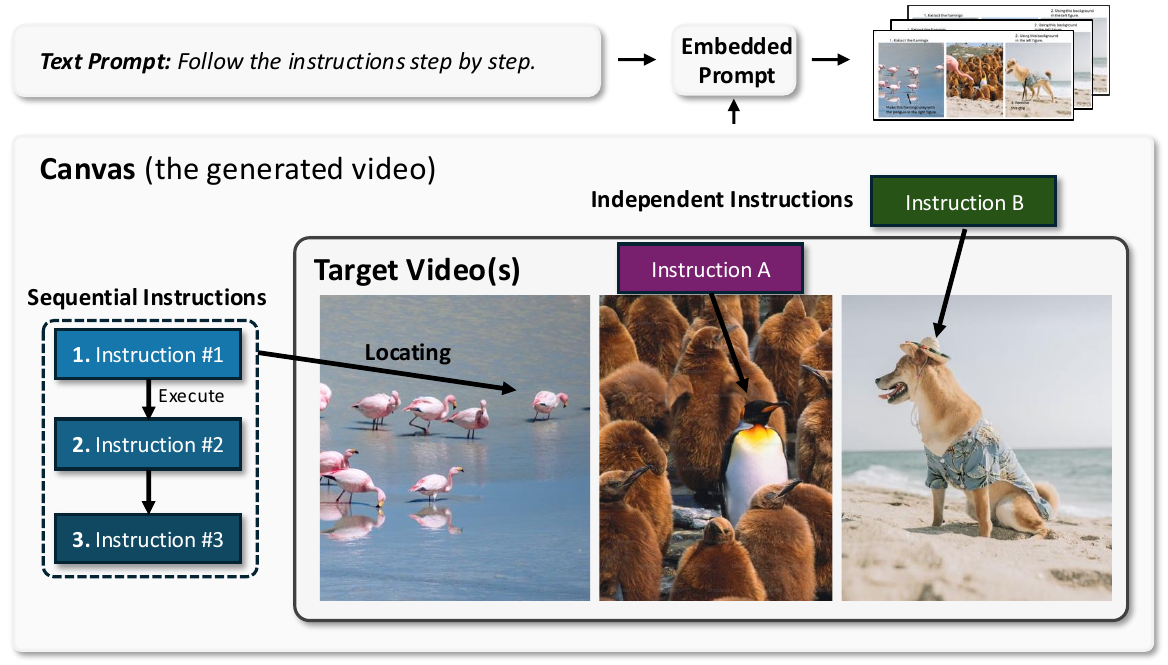}
    \caption{In-Video Instruction controls generation by placing the instruction directly on the first frame, providing explicit spatial grounding for the instruction’s scope. This enables assigning independent, less ambiguous, and even multi-step sequential commands to different targets. During generation, we fix the textual prompt to “follow the instructions step by step” and rely solely on in-frame visual signals for control.}
    \label{fig:framework}
\end{figure}

The construction of In-Video Instructions begins with an initial video frame containing one or more objects of interest. Human guidance is embedded into this frame through minimal visual annotations. In this work, we instantiate the instruction space using two basic primitives:
\begin{itemize}
    \item \textbf{Short textual commands}, which specify the intended behavior of a subject;
    \item \textbf{Arrows}, which serve as spatial indicators that localize the target and may also convey the direction or region of influence.
\end{itemize}
As illustrated in Figure~\ref{fig:framework}, multiple instructions may coexist within the same frame, forming either one-to-one or one-to-many correspondences between subjects and commands. Instruction placement is flexible: text can be positioned within the target region to establish a direct association, or placed externally to preserve a clear video scene. Directional elements such as arrows can be added to indicate the relevant subject or region and further strengthen the spatial linkage. In addition, the method naturally supports multi-step instructions by assigning explicit ordering to each command, for instance using numbered labels such as ``1. Instruction \#1'' and ``2. Instruction \#2.''

\begin{figure*}[t]
    \centering
    \includegraphics[width=\linewidth]{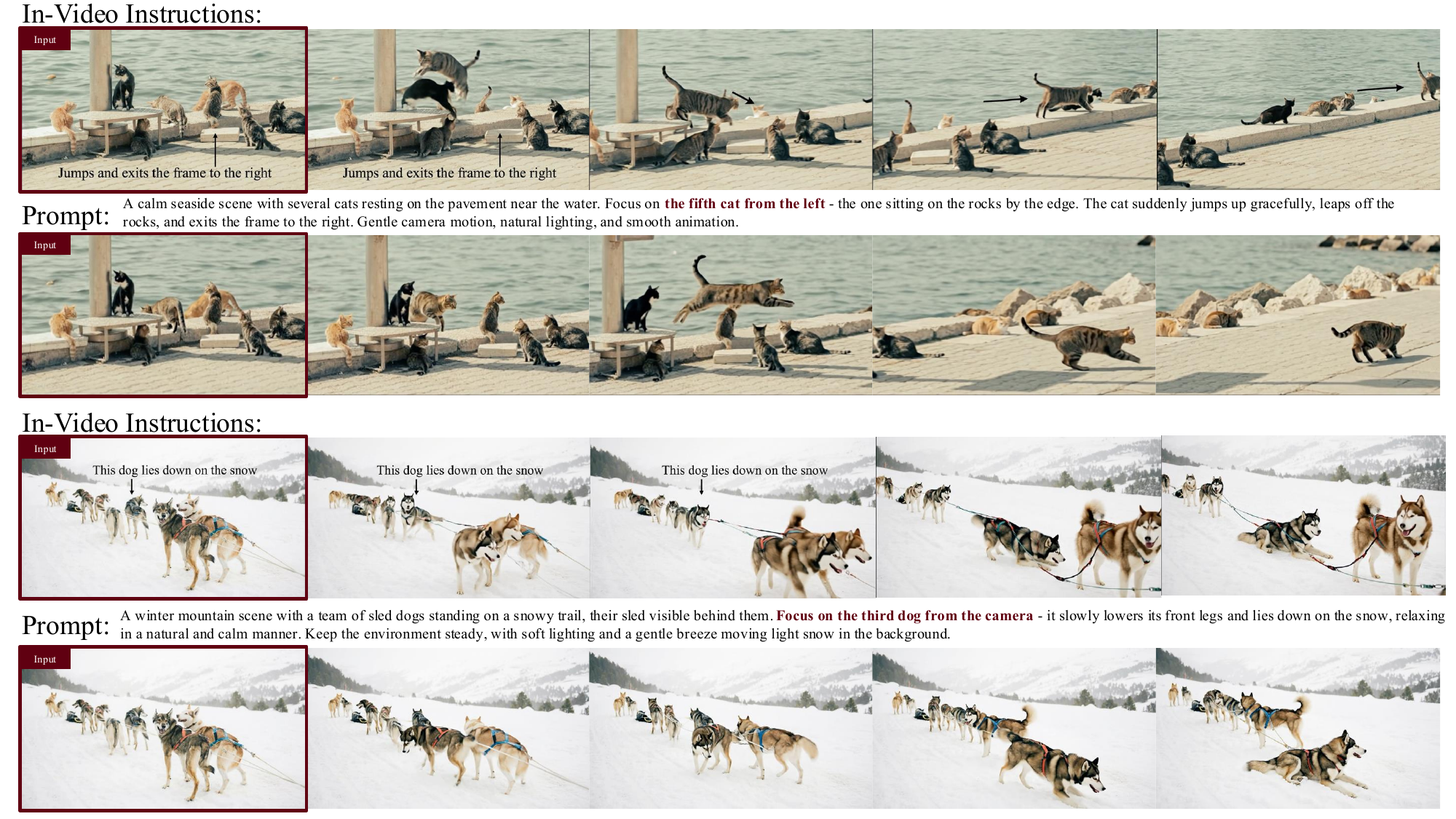}
    \caption{Spatial Localization Ability of In-Video Instructions. We use In-Video Instructions to localize a target object among multiple entities and execute the corresponding action. For the prompt-based baseline, we rely on ChatGPT-generated textual descriptions such as ``the N-th object from the left'' for locating. As shown, In-Video Instructions enable precise and unambiguous localization, whereas text-only prompts exhibit noticeable limitations in resolving object positions.}
    \label{fig:location}
\end{figure*}

\subsection{Generation Procedure}

Given an annotated frame, the generation process follows the standard inference pipeline of a pretrained video generative model in the image-to-video setting. The annotated frame is supplied as the initial conditioning frame, and a single global text prompt is used to reinforce adherence to the visual instructions:
\begin{quote}
\texttt{\textbf{Fixed Text Prompt:} Follow the instructions step by step.}
\end{quote}
The text prompt is optional and can be removed or embedded directly within the canvas for unified control (See Figure \ref{fig:multiple_frames}). No finetuning or architectural modification is applied. During inference, the model interprets the overlaid text and arrows as integral components of the input scene and implicitly treats them as actionable signals. The instructions appear only in the first frame, while subsequent frames are synthesized freely by the model, which propagates the intended motion, pose change, or interaction over time. In practice, this simple protocol is sufficient to induce a broad range of controllable behaviors, including localized object motion, camera movement, and multi-step or multi-object actions, as demonstrated in our experiments.

\section{Experimental Results}

In this section, we empirically investigate the performance of In-Video Instruction across diverse and complex scenarios, providing both quantitative and qualitative analyses. Our experiments are conducted on both commercial models, including Veo-3.1~\cite{veo2025} and Kling-2.5~\cite{Kuaishou2024KlingAI}, as well as open-source models such as WAN-2.2~\cite{wan2025wan}. Unless stated otherwise, videos are generated with Veo-3.1 by default. We first focus on two fundamental and essential capabilities of In-Video Instruction: (1) the ability to comprehend textual commands embedded within the image and execute the corresponding actions, and (2) the capability to locate and interact with specific visual subjects, thereby enabling fine-grained and less ambiguous generation control. Beyond these basic abilities, we further explore In-Video Instruction’s potential for controlling object and camera motion, as well as its effectiveness in multi-object and complex interaction scenarios.

\begin{table*}[t]
    \centering
    \begin{tabular}{l | cc | cc | cc}
    \toprule
        \multirow{2}{*}{\bf Dimensions} 
        & \multicolumn{2}{c|}{\bf Veo3.1 Fast (16:9, 720P)} 
        & \multicolumn{2}{c|}{\bf Kling-2.5 (1:1, 720P)} 
        & \multicolumn{2}{c}{\bf Wan2.2-A13B (1:1, 480P)} \\
        \cmidrule(lr){2-3} \cmidrule(lr){4-5} \cmidrule(lr){6-7}
        & In-Video Inst. & Text Prompt 
        & In-Video Inst. & Text Prompt 
        & In-Video Inst. & Text Prompt \\
        \midrule
        Subject & 0.9710 & \bf 0.9842 & 0.9824 & \bf 0.9933 & 0.9823 & \bf 0.9861 \\
        Dynamic Degree & \bf 0.8392 & 0.7857 & \bf 0.5625 & 0.4218 & \bf 0.5859 & 0.4921 \\
        Motion Smoothness & \bf 0.9911 & 0.9907 & 0.9905  & \bf 0.9911 & 0.9791 & \bf 0.9799 \\
        Aesthetic Quality & 0.5943 & 0.\bf 6006 & 0.6553 & \bf 0.6644 & 0.6173 & \bf 0.6336 \\
        Imaging Quality & 0.7074  & \bf 0.7086 & 0.7440 & \bf 0.7507 & \bf 0.7229 & 0.7221 \\
        Temporal Flickering & 0.9710 & \bf 0.9719 & 0.9664 & \bf 0.9793 & \bf 0.9633 & 0.9632 \\
    \bottomrule
    \end{tabular}
    \caption{VBench evaluation of videos generated with in-video instructions and traditional text prompts. For in-video instructions, the prompt text is directly embedded at the top of each frame. Results show that understanding and following in-video instructions remains more challenging than responding to text prompts. Due to resolution constraints, we use a 16:9 ratio for Veo3.1 and 1:1 for other models.}
    \label{tab:vbench_invideo_text}
\end{table*}

\begin{figure*}[t]
    \centering
    \includegraphics[width=\linewidth]{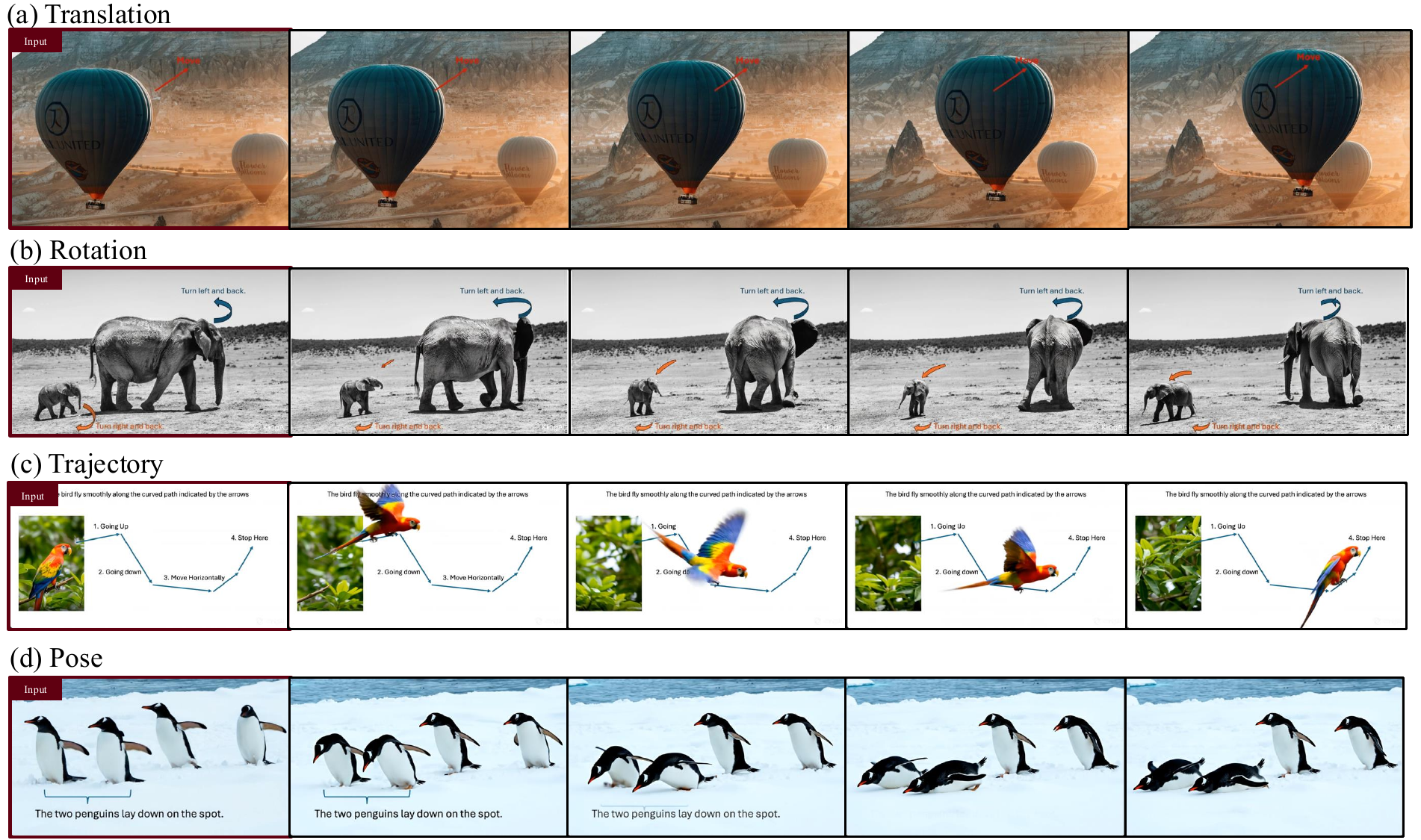}
    \caption{Controlling object motions or trajectories with in-video instructions.}
    \label{fig:motion}
\end{figure*}

\subsection{Text Understanding and Locating}

\begin{figure*}[t]
    \centering
    \includegraphics[width=\linewidth]{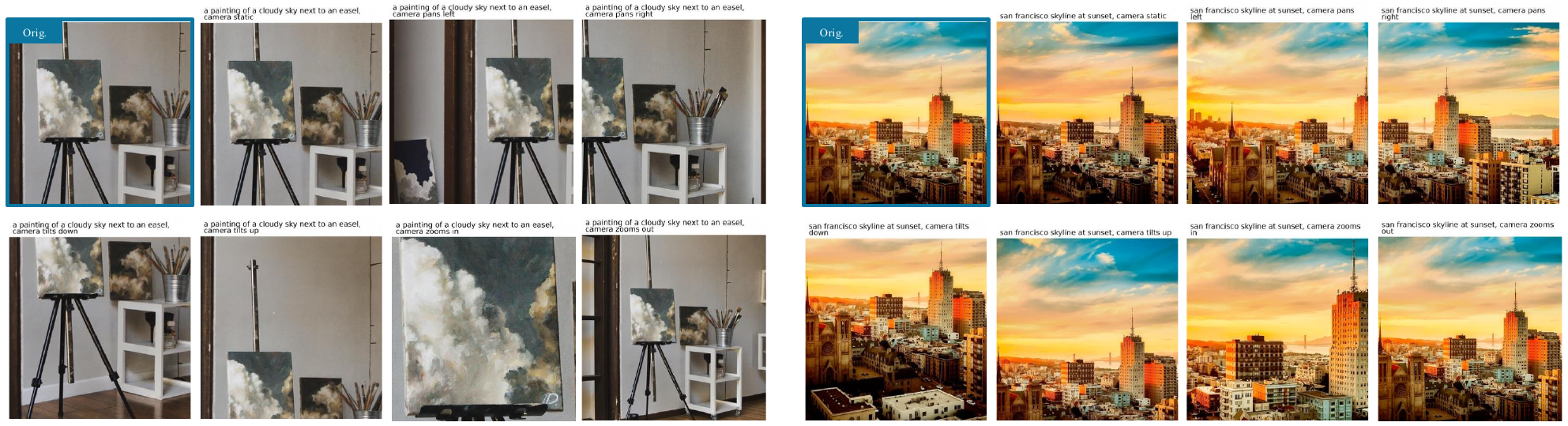}
    \caption{Controlling camera motion with In-Video Instructions. We visualize the initial frame and the final output for seven camera-motion types: static, pan left, pan right, tilt down, tilt up, zoom in, and zoom out.}
    \label{fig:camera_motion}
\end{figure*}

\paragraph{The ability to understand text instructions.} Text input is one of the most common forms of control in video generation, as it effectively specifies high-level objectives such as desired content, object motion, and scene transitions. We first demonstrate that multimodal inputs in image-to-video models can obtain generative signals not only from textual prompts, but also from text embedded within the visual input itself. To validate this, we evaluate Veo-3.1, Kling-2.5, and Wan-2.2 on the VBench benchmark~\cite{huang2024vbench,huang2024vbench++} as shown in Table~\ref{tab:vbench_invideo_text}. Each input in this task consists of a textual prompt paired with an initial frame. We compare the performance of In-Video Instruction with that of conventional text prompts. For the In-Video Instruction setting, we embed the textual command as a caption above the image and feed the combined image into the model, and fix the input prompt as ``Follow the instructions step by step.'' Some examples are shown in Figure~\ref{fig:camera_motion}. In the VBench evaluation, we observe that In-Video Instruction performs slightly below but remains close to the performance of direct text inputs, demonstrating the model’s basic ability to interpret text embedded in images. The mild performance gap is expected, as interpreting text from visual content is naturally more challenging than processing explicitly provided textual prompts. 

\paragraph{Spatial Locating and Interaction.} Compared with purely understanding text, the unique strength of In-Video Instruction lies in its capability for spatial localization and interaction. While conventional text prompts effectively convey global semantics, they often lack fine-grained control over local regions, making it difficult to direct individual object behaviors in multi-object scenes. In-Video Instruction overcomes this limitation by allowing spatially placed textual commands and visual markers (e.g., arrows) to specify distinct actions for different objects, enabling precise and interpretable control within complex visual environments and dynamic interactions. To validate this ability, we qualitatively examine its instruction localization in multi-object scenarios. As shown in Figure~\ref{fig:location}, we compare results from traditional text prompts and our approach, using ChatGPT to automatically generate unbiased spatial expressions such as “the N-th object from the left” to avoid human-crafted phrasing biases. Results show that In-Video Instruction achieves accurate object–instruction alignment and controllable generation in multi-object scenes, where text-only prompts often struggle due to spatial ambiguity. This demonstrates that visual grounding provides a powerful mechanism for assigning explicit behavioral directives to different entities. Moreover, it highlights the potential of In-Video Instruction as a flexible and expressive interface for interpretable and fine-grained video control.

\subsection{Motion Control} 

Building upon the model’s fundamental abilities in text understanding and spatial interaction, we further examine motion control, a central aspect of controllable video generation that governs both object and camera dynamics. As illustrated in Figure \ref{fig:motion}, we categorize four representative motion types: translation, rotation, trajectory, and pose, each defined by visual signals such as arrows, curves, or short textual annotations embedded in the first frame. These cues provide direct and interpretable spatial conditioning and enable precise, fine-grained motion guidance.

\paragraph{Translation.} The key challenge in motion control primarily lies in direction specification, which is often coarse and imprecise in text-based descriptions. In-Video Instruction addresses this limitation by anchoring visual arrows directly to the target object, enabling the model to infer both the motion vector and its magnitude. As illustrated by the hot-air balloon example in Figure~\ref{fig:motion}(a), the object’s movement aligns precisely with the intended initial direction.

\paragraph{Rotation.} Rotation is comparatively easier to express, yet still benefits from explicit visual grounding. Curved arrows intuitively convey rotational direction and pivot centers, allowing the model to perform controlled rotation. In the elephant example in Figure~\ref{fig:motion}(b), different rotation instructions are assigned to different objects, resulting in independent yet coordinated rotations.

\paragraph{Trajectory.} Trajectory control represents a more complex form of motion, requiring the model to follow multi-stage or curved paths. While describing such motion textually (e.g., “fly upward, then turn left, and stop”) is cumbersome and ambiguous, In-Video Instruction allows users to directly sketch trajectories as continuous curves. The model accurately follows these drawn paths, maintaining realistic dynamics and temporal consistency throughout the sequence.

\paragraph{Pose.} In-Video Instruction enables coherent and smooth pose adjustments, allowing the model to generate natural and consistent variations in posture. This demonstrates that the model can effectively interpret localized visual and textual information as actionable control signals, achieving fine-grained pose manipulation.

\begin{figure*}[t]
    \centering
    \includegraphics[width=\linewidth]{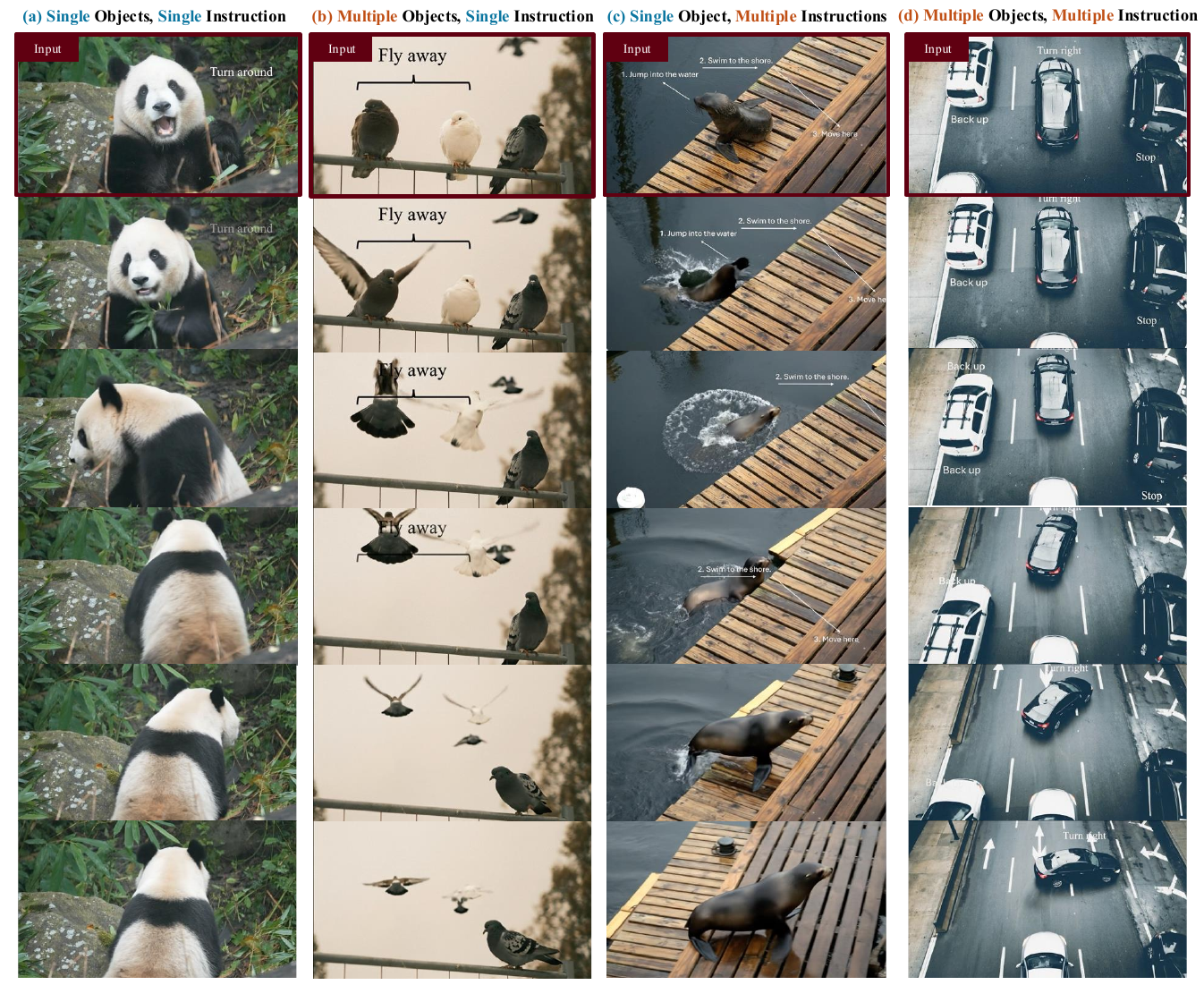}
    \caption{In-Video Instructions with Multiple Objects and Commands, enabling both sequential instructions that involve a series of actions and parallel instructions that manipulate different objects independently.}
    \label{fig:multiple_objects}
\end{figure*}

\paragraph{Camera Motion.} Beyond object motion, another key aspect of video dynamics lies in camera movement. To examine whether text embedded within an image can control camera motion, we adopt the same evaluation strategy as in Table~\ref{tab:vbench_invideo_text}, embedding camera-related commands as captions above the image input. This setup allows us to assess the capability of In-Video Instruction in controlling camera movement. Figure \ref{fig:camera_motion} illustrates seven distinct camera motions: static, pan left, pan right, tilt down, tilt up, zoom in, and zoom out. As shown, textual cues within the image can effectively guide camera behavior. 

\paragraph{When to Use In-Video Instructions.} In summary, In-Video Instruction is capable of controlling both object motion and camera motion. Among them, object motion control is inherently more localized and, in complex multi-object scenes, highly dependent on accurate spatial localization. In contrast, camera motion represents a more global form of control that affects the entire scene. From this perspective, In-Video Instruction is particularly well-suited for fine-grained and spatially grounded manipulation. However, we note that this capability is not essential for camera motion, as it represents a more global form of control that can be well handled by conventional text prompts. 

\subsection{Multiple Objects and Instructions}

We further evaluate the scalability of In-Video Instruction by examining its performance across configurations involving multiple instructions, objects, and their combinations. Specifically, we study four representative scenarios:
(1) single instruction, single object, where one subject responds to a single command;
(2) single instruction, multiple objects, where several subjects execute the same action concurrently;
(3) multiple instructions, single object, where a single subject performs a sequence of ordered actions; and
(4) multiple instructions, multiple objects, where multiple entities each follow distinct and independent instructions.
These settings allow us to systematically analyze the performance of In-Video Instructions across different scenes and compare their advantages over conventional text prompts.

\paragraph{Single Object, Single Instruction.}
This serves as the simplest and most direct control setting, validating the model’s ability to ground a single instruction. As shown in Figure~\ref{fig:multiple_objects}(a), when prompted to “turn around,” the panda executes a smooth and coherent rotation, demonstrating that both text-prompt and in-video instruction settings achieve comparable performance. This suggests that for simple, globally interpretable tasks, textual and visual instructions are equally effective.

\paragraph{Multiple Objects, Single Instruction.}
When a shared instruction applies to multiple entities, the ability to specify spatial correspondence becomes more relevant. As illustrated in Figure~\ref{fig:multiple_objects}(b), two birds respond to the “fly away” instruction while the third remains stationary. The spatial anchoring of the instruction helps the model associate the action with the intended subjects, ensuring consistent yet selective control. This demonstrates that visual grounding offers a practical mechanism for managing concurrent behaviors among multiple entities within the same scene.

\paragraph{Single Object, Multiple Instructions.}
The next scenario requires sequential reasoning, where one subject performs a series of temporally dependent actions. As illustrated in Figure \ref{fig:multiple_objects}(c), the seal follows three ordered commands: jump into the water, swim to the shore, and move here, forming a continuous motion with correct temporal logic and spatial alignment. In-Video Instructions encode stepwise relations directly within the visual domain, where the spatial ordering and numbering of cues provide implicit temporal structure. Compared to text prompts, In-Video Instructions make it easier for the model to handle interactions between objects and their environment, combining spatial localization and trajectory reasoning to construct complex and controllable video generations.

\paragraph{Multiple Objects, Multiple Instructions.}
The most complex scenario involves issuing distinct instructions to multiple entities within a single frame. In Figure~\ref{fig:multiple_objects}(d), three cars perform different actions such as backing up, turning right, and stopping, while preserving coherent scene dynamics. Our result demonstrates that the model can interpret spatially separated visual signals as independent control signals, enabling localized manipulation without mutual interference. In contrast to text prompts, which express only global intent with limited positional specificity, In-Video Instructions offer flexible, target-aware control and produce precise, disentangled behaviors. 

\paragraph{The success rate of multiple instructions.} To further evaluate the effectiveness, we conducted a human assessment comparing In-Video Instructions with conventional text-based prompting on the ``Multiple Objects, Multiple Instructions'' setting in Figure~\ref{fig:multiple_objects}(d), generating 24 videos for each method and evaluating them through human judgment. As shown in Table~\ref{tab:success_rate}, the model consistently follows the embedded visual instructions in this complex setting, achieving higher success rates than text-only prompts across diverse motion patterns. In addition, we observed an interesting phenomenon: instructing the white car in Figure~\ref{fig:multiple_objects}(d) to back up is relatively difficult, which leads to a success rate of 20.8\% with the in-video instruction and 8.3\% with the text prompt. This appears to stem from the presence of another vehicle directly behind it, which induces a strong prior for the model to move the car forward rather than backward, particularly when resolving physically plausible trajectories.

\begin{table}[t]
    \centering
    \small
    \begin{tabular}{l |c c}
    \toprule
        \bf  & \bf In-Video Inst. & \bf Prompt \\
        \midrule
        Instruction A (Back up) & \bf 20.8\% & 8.3\%  \\  
        Instruction B (Turn right) & \bf 58.3\% & 29.2\%  \\ 
        Instruction C (Stop) & \bf 95.8\% & 58.3\% \\ 
    \bottomrule
    \end{tabular}
    \caption{Success rates of instructions under the ``multiple objects, multiple instructions'' setting in Figure~\ref{fig:multiple_objects}(d), averaged over 24 generated videos based on human evaluation.}
    \label{tab:success_rate}
\end{table}

\begin{figure}[t!]
    \centering
    \includegraphics[width=\linewidth]{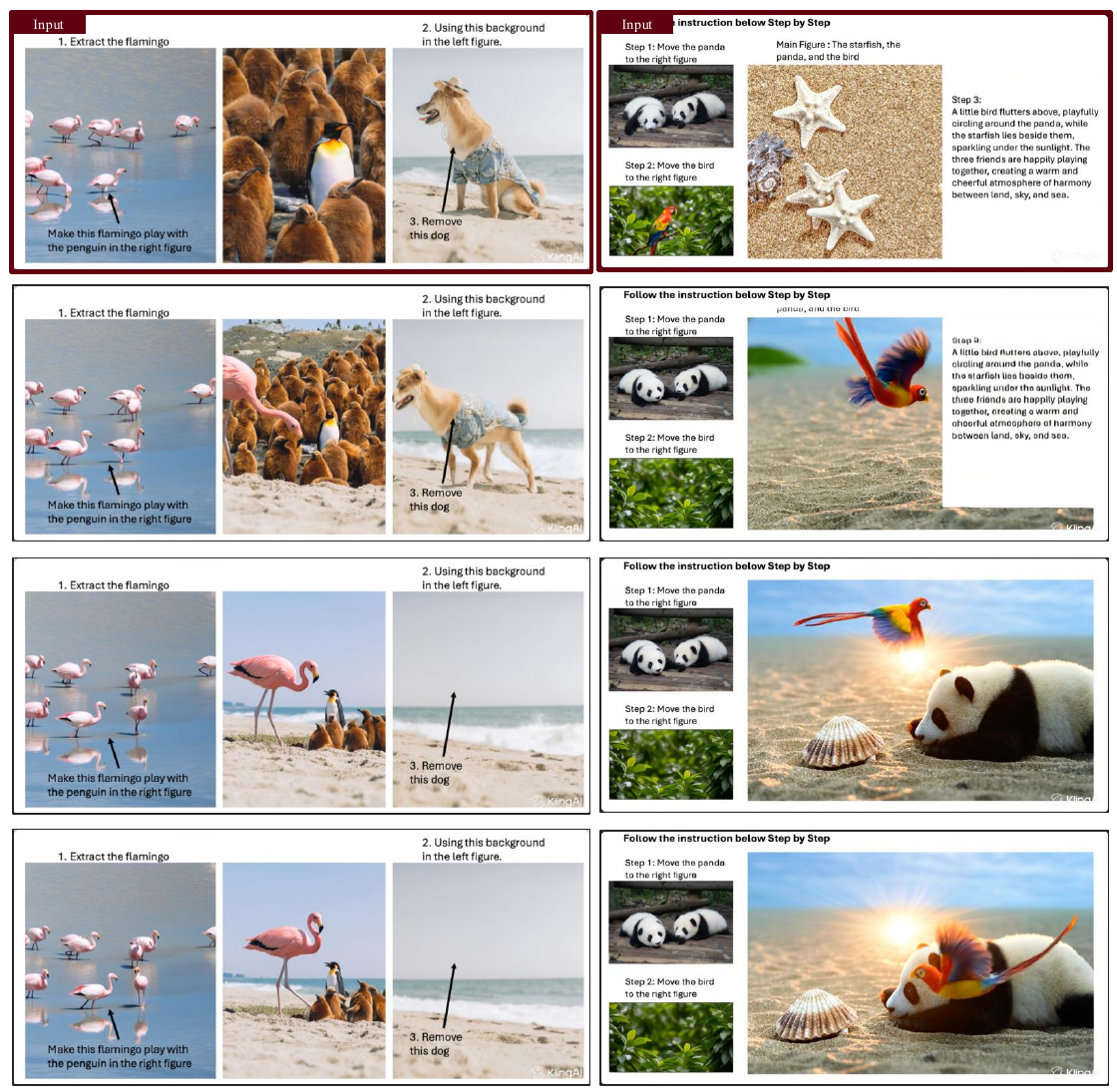}
    \caption{Synthesizing videos by manipulating multiple seed frames. We generate videos from several initial frames and use visual instructions to coordinate interactions across them; all videos in this setting are produced using Kling-2.5.}
    \label{fig:multiple_frames}
\end{figure}

\paragraph{Synthesizing Videos by Manipulating Multiple Frames.} Beyond single-frame control, we show that advanced video models can synthesize complex scenes by integrating information from multiple source frames. As shown in Figure~\ref{fig:multiple_frames}, the model combines spatial and temporal cues from different visual inputs to produce coherent and continuous video sequences. This demonstrates the ability to interpret cross-frame instructions and maintain consistency across distinct visual contexts. 

\section{Limitations}
While In-Video Instruction offers a simple and intuitive way to guide generation, several limitations remain. Since the instructions are drawn directly on the image, they persist in the generated video and often require post-processing for removal. We also observe that these visual markers may become occluded during synthesis, suggesting that the model already possesses priors for suppressing such elements. Extending the text prompt to explicitly remove visible annotations may therefore further improve the results. In addition, our analysis remains largely qualitative, underscoring the need for more systematic assessment in future work. Finally, all instructions examined in this study are manually constructed, whereas real-world videos contain inherent visual signals such as traffic lights or signboards; understanding whether models can interpret and react to these natural signals remains an interesting direction for future research.

\section{Conclusion}

This work introduces \emph{In-Video Instruction}, a simple and training-free approach that embeds human intent directly into the visual input for controllable video generation. The method enables fine-grained, spatially grounded control across diverse tasks, and experiments on multiple models demonstrate that these visual signals can be reliably interpreted as actionable guidance, offering strong flexibility and controllability in complex scenes.

{
    \small
    \bibliographystyle{ieeenat_fullname}
    \bibliography{main}
}


\end{document}

%% file: preamble.tex
\usepackage{graphicx}
\usepackage{amsmath}
\usepackage{amssymb}
\usepackage{booktabs}

\usepackage{multirow}
\usepackage{wasysym}

\usepackage{makecell}
\usepackage{lipsum}
\usepackage{subcaption}
\usepackage{amsmath}
\usepackage{amssymb}
\usepackage{booktabs}
\usepackage{multirow}
\usepackage{multicol}
\usepackage{bbm}
\usepackage{lipsum}
\usepackage{colortbl}
\usepackage{amsthm}
\usepackage{graphicx}
\usepackage{wrapfig}
\usepackage[ruled, linesnumbered, noend]{algorithm2e}